%% file: cvpr_21_avi_v7.tex
\documentclass[final]{cvpr}

\usepackage{graphicx}
\usepackage{footmisc}
\usepackage{booktabs}       
\usepackage{amsfonts}       
\usepackage{nicefrac}       
\usepackage{microtype}      
\usepackage{color}
\usepackage{bm}
\usepackage{amsmath}
\usepackage{wrapfig}
\usepackage{graphicx}
\usepackage{setspace}
\usepackage[super]{nth}
\usepackage{siunitx}

\usepackage{caption}
\def\cvprPaperID{1500}
\usepackage{blindtext}
\usepackage{marvosym}
\usepackage{times}
\usepackage{epsfig}
\usepackage{graphicx}
\usepackage{amsmath}
\usepackage{amssymb}
\usepackage{colortbl}
\definecolor{mygray}{gray}{.88}

\newcommand{\XXX}{SGM}

\usepackage[pagebackref=true,breaklinks=true,colorlinks,bookmarks=false]{hyperref}
\usepackage{hyperref}
\usepackage{bm}

\usepackage{caption}
\usepackage[caption=false]{subfig}
\usepackage{multirow}
\usepackage{multicol}
\usepackage{adjustbox}

\usepackage{bm}

\def\confYear{CVPR 2021}

\begin{document}

\pagenumbering{gobble}

\title{Deep Animation Video Interpolation in the Wild}

\author{{Li Siyao$^{1*}$} $\,\,\,\,\,\,\,$ Shiyu Zhao$^{1,2*}$ $\,\,\,\,\,\,\,$ Weijiang Yu$^{3}$ $\,\,\,\,\,\,\,$ Wenxiu Sun$^{1,4}$ \\  Dimitris Metaxas$^{2}$ $\,\,\,\,\,\,\,\,$ Chen Change Loy$^{5}$ $\,\,\,\,\,\,\,\,$ Ziwei Liu\textsuperscript{5~\Letter}\\

$^{1}$SenseTime Research and Tetras.AI $\,\,\,\,\,\,$  $^{2}$Rutgers University $\,\,\,\,\,\,$ $^{3}$Sun Yat-sen University  \\ $^{4}$Shanghai AI Laboratory $\,\,\,\,\,$ $^{5}$S-Lab, Nanyang Technological University\\

{\tt\small lisiyao1@sensetime.com} $\,\,\,$ {\tt\small sz553@rutgers.edu} $\,\,\,$ {\tt\small weijiangyu8@gmail.com} \\ {\tt\small sunwenxiu@sensetime.com} $\,\,\,$
 {\tt\small dnm@cs.rutgers.edu} $\,\,\,$ {\tt\small \{ccloy, ziwei.liu\}@ntu.edu.sg} 
%

}

\vspace{-40pt}
\twocolumn[{%
\renewcommand\twocolumn[1][]{#1}%
\maketitle
\begin{center}
    \centering
    \small{
     \vspace{-15pt}
    \includegraphics[width=0.95\linewidth]{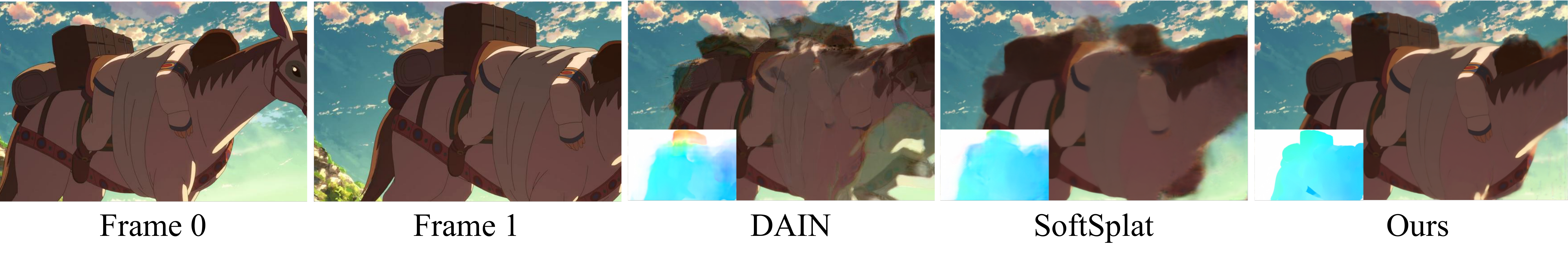}
        
        


        }
    \vspace{-12pt}
    \captionof{figure}{\textbf{A typical example for animation video interpolation.} Our approach is capable of estimating optical flows for large motions correctly and restore the content, while competing methods fail to handle such motions.} 
    \label{fig:teaser}
\end{center}%
}]
\maketitle

\vspace{-7pt}
\begin{abstract}
\vspace{-7pt}
In the animation industry,{\let\thefootnote\relax\footnote{{$^{*}$Equal contributions; \textsuperscript{\Letter}\text{Corresponding author}.}}} cartoon videos are usually produced at low frame rate since hand drawing of such frames is costly and time-consuming.
Therefore, it is desirable to develop computational models that can automatically interpolate the in-between animation frames.
However, existing video interpolation methods fail to produce satisfying results on animation data. 
Compared to natural videos, animation videos possess two unique characteristics that make frame interpolation difficult: 1) cartoons comprise lines and smooth color pieces. 
The smooth areas lack textures and make it difficult to estimate accurate motions on animation videos.
2) cartoons express stories via exaggeration. Some of the motions are non-linear and extremely large.
In this work, we formally define and study the animation video interpolation problem for the first time.
To address the aforementioned challenges, we propose an effective framework, \textbf{AnimeInterp}, with two dedicated modules in a coarse-to-fine manner.
Specifically, 1) Segment-Guided Matching resolves the ``lack of textures'' challenge by exploiting global matching among color pieces that are piece-wise coherent.
2) Recurrent Flow Refinement resolves the ``non-linear and extremely large motion'' challenge by recurrent predictions using a transformer-like architecture. 
To facilitate comprehensive training and evaluations, we build a large-scale animation triplet dataset, \textbf{ATD-12K}, which comprises 12,000 triplets with rich annotations.
Extensive experiments demonstrate that our approach outperforms existing state-of-the-art interpolation methods for animation videos.
Notably, AnimeInterp shows favorable perceptual quality and robustness for animation scenarios in the wild.
The proposed dataset and code are available at \url{https://github.com/lisiyao21/AnimeInterp/}.

\end{abstract}

\vspace{-16pt}
\section{Introduction}
\label{sec:intro}
\input{content/intro}


\begin{figure*}	
	\centering
\includegraphics[width=0.95\linewidth]{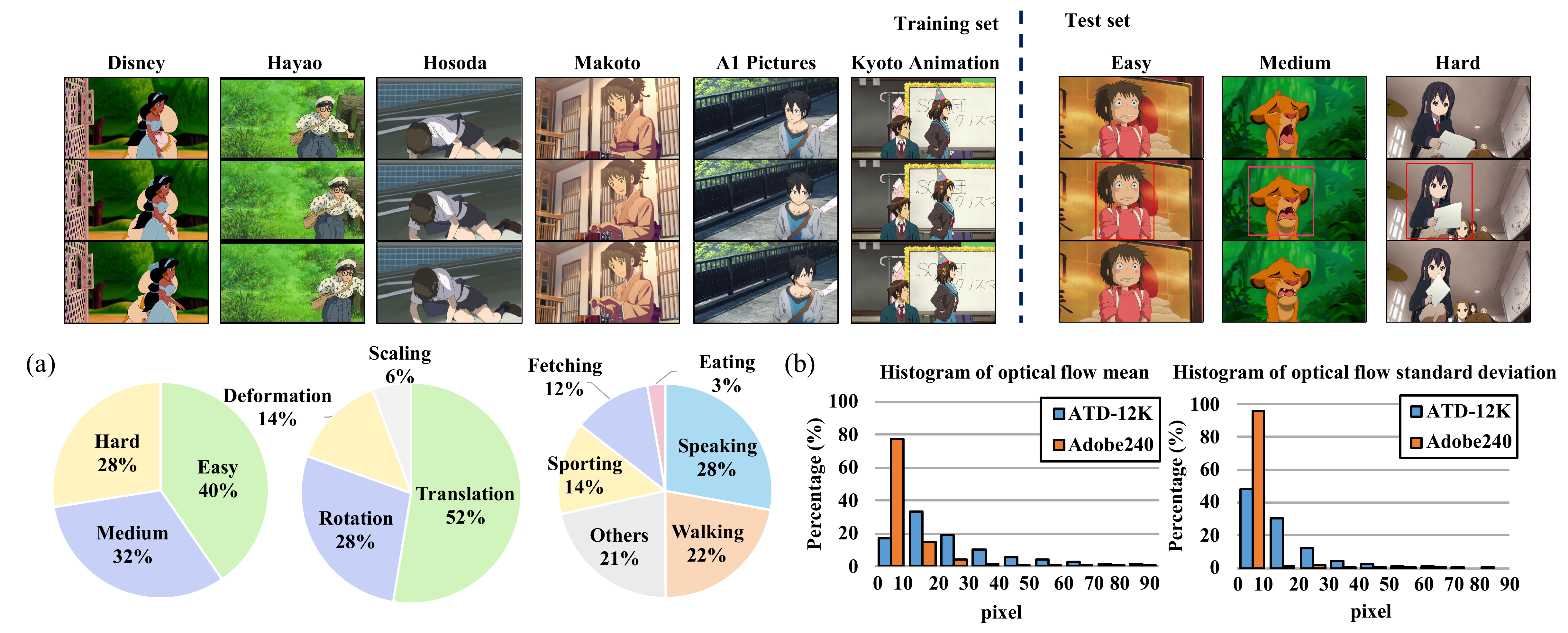}
\vspace{-10pt}
\caption{\textbf{Triplet samples and statistics of ATD-12K.} Top: typical triplets in the training and test sets. (a) The percentage of difficulty levels and motion tags in different categories. (b) Histograms of mean motion values and standard deviations calculated in each image.}\label{fig:ani_triplets}
\vspace{-12pt}
\end{figure*}

\vspace{-6pt}
\section{Related Work}
\label{sec:related}
\input{content/related}


\section{ATD-12K Dataset}
\label{sec:dataset}

\input{content/dataset}


\section{Our Approach}
\label{sec:algorithm}

\input{content/algorithm}

\section{Experiments}
\label{sec:experiment}
\input{content/experiment}

\section{Conclusion}
\label{sec:conclusion}
\input{content/conclusion}

 {\noindent\textbf{Acknowledgement.}}
{This research was conducted  in collaboration with SenseTime. This work is supported by NTU NAP and A*STAR through the Industry Alignment Fund - Industry Collaboration Projects Grant.
}

{\small
\bibliographystyle{ieee_fullname}
\bibliography{egbib.bib}
}

\end{document}


\title{Deep Animation Video Interpolation in the Wild \\ Supplementary File}

\author{{Li Siyao$^{1*}$} $\,\,\,\,\,\,\,$ Shiyu Zhao$^{1,2*}$ $\,\,\,\,\,\,\,$ Weijiang Yu$^{3}$ $\,\,\,\,\,\,\,$ Wenxiu Sun$^{1,4}$ \\  Dimitris Metaxas$^{2}$ $\,\,\,\,\,\,\,\,$ Chen Change Loy$^{5}$ $\,\,\,\,\,\,\,\,$ Ziwei Liu\textsuperscript{5~\Letter}\\

$^{1}$SenseTime Research and Tetras.AI $\,\,\,\,\,\,$  $^{2}$Rutgers University $\,\,\,\,\,\,$ $^{3}$Sun Yat-sen University  \\ $^{4}$Shanghai AI Laboratory $\,\,\,\,\,$ $^{5}$S-Lab, Nanyang Technological University\\

{\tt\small lisiyao1@sensetime.com} $\,\,\,$ {\tt\small sz553@rutgers.edu} $\,\,\,$ {\tt\small weijiangyu8@gmail.com} \\ {\tt\small sunwenxiu@sensetime.com} $\,\,\,$
 {\tt\small dnm@cs.rutgers.edu} $\,\,\,$ {\tt\small \{ccloy, ziwei.liu\}@ntu.edu.sg} }

\twocolumn[{%
\renewcommand\twocolumn[1][]{#1}%
\maketitle

}]
\maketitle

\begin{abstract}
     In this supplementary file, we provide some details of the dataset {\bf{ATD-12K}} and illustrate network architectures of the proposed method {\bf{AnimeInterp}}. 
     Moreover, with the motion tags of ATD-12K, we benchmark the performance of various methods on different motion categories.
     A demo video, which contains several clips of interpolation results generated by {\bf{AnimeInterp}} and other state-of-the-art methods, is attached to further demonstrate the effectiveness of our method on animation video interpolation.

\end{abstract}

\section{Details of ATD-12K}

\noindent{\bf Data sources} (Section 3.1 in the main paper).
Triplets of ATD-12K are collected from 30 cartoon movies made by diversified producers.
The training set is collected from 22 movies, and the test set is collected from the rest 8.
The detailed movie names are listed as follows:
\begin{itemize}
\vspace{-2mm}
\setlength{\itemsep}{0pt}
\setlength{\parsep}{0pt}
\setlength{\parskip}{0pt}
\item {\bf Movies for the training set:} \it{Brother Bear}; Treasure Planet; Atlantis: The Lost Empire;  Lilo \& Stitch; The Emperor's New Groove; Tarzan; Mulan; The Hunchback of Notre Dame; Pocahontas; Beauty and the Beast; Aladdin; The Little Mermaid; The Adventures of Ichabod and Mr. Toad;  Princess Mononoke;  Howl's Moving Castle; The Wind Rises; Your Name; Time Traveller: The Girl Who Leapt Through Time; The Disappearance of Haruhi Suzumiya; Sword Art Online The Movie Ordinal Scale; Cinderella; A Silent Voice.
\item {\bf Movies for the test set:} The Lion King; The Jungle Book; Peter Pan; Alice in Wonderland; Spirited Away; Children who Chase Lost Voices; The Boy and The Beast; K-ON!
\end{itemize}

\noindent{\bf Difficulty level} (Section 3.2).
We divided triplets of the ATD-2K into three levels, \ie, ``Easy'', ``Medium'', and ``Hard'', which indicate the difficulty to generate the middle image $I_{1/2}$ of a triplet with $I_0$ and $I_1$.
%
%
Specifically, the difficulty levels are labeled based on the average magnitude of motion and the the ratio of the occlusion area in the input frame.
First, we estimate the optical flow $f_{0\to1}$ of $I_0$ and $I_1$. 
Then, we compute the mean $\bar f_{0\to1}$ and standard deviation $\sigma_{f_{0\to1}}$ of the magnitude of $f_{0\to1}$. 
Next, we use $f_{0\rightarrow1}$ to splat a tensor (denoted as $\mathbf 1$), which has the same size of $I_0$ and is filled with 1.
The splatting result is notated as $\mathbf 1_{0\rightarrow1}$.
In $1_{0\rightarrow1}$, pixels with values smaller than 0.05 is regarded to be occluded, and the occlusion area rate $O_{f_{0\to1}}$ is computed as the ratio of the number of the occluded pixels and the area of the whole image.
Based on calculated $\bar f_{0\to1}$, $\sigma_{f_{0\to1}}$, and $O_{f_{0\to1}}$, we define the difficulty levels according to the rules shown in Table~\ref{table:division_level}. 
Note that we take  not only  the mean magnitude  but also the inner variation of motion into judgement to reflect the complexity  more objectively. 

\begin{table} [t]
\renewcommand\arraystretch{1.2}
    \centering
    \caption{{\bf Definition of difficulty levels.}}\label{table:division_level}
    \resizebox{.95\columnwidth}{!}{
    \begin{tabular}{c|ccc}
    \bottomrule
       \multirow{2}{*}{} & \multicolumn{3}{c}{Range of $O_{f_{0\to1}}$} \cr
      \cline{2-4}
     & $[0, 0.05)$ & $[0.05, 0.2)$ & $[0.2, \infty)$ \cr
    \hline
     $\bar f_{0\to1} > 10$, $\sigma_{f_{0\to1}} > 10$ & Middle & Hard &Hard \cr
    
    $\bar f_{0\to1} > 10$, $\sigma_{f_{0\to1}} \leq 10$ & Easy & Middle & Hard \cr
    
    $\bar f_{0\to1} \leq 10$ & Easy & Easy & Middle \cr
    \toprule
    \end{tabular}
    }
\end{table}

\begin{figure*}	
	\centering
	\includegraphics[width=1\linewidth, trim= 0 50pt 0 0pt, clip]{figures/supp_fig_cr.pdf}
\caption{{\bf Evaluations on different motion tags of ATD-12K.} AnimeInterp achieves the leading performance on all tags.}\label{fig:motion_tags}
\end{figure*}

\section{Details of AnimeInterp}

\noindent{\bf Contour Extraction and Color Piece Segmentation} (Section 4.1).
Unlike edges of real image, the contours of a cartoon are explicitly drawn with non-negligible width.
If one use a regular edge detection algorithms (\eg Canny detector) on anime image, two seperative lines will be produced for both sides of the contour, which will yield wrong segmentation in the next step. 
Instead, we adopt convolution on the image with a $5\times 5$ Laplacian kernel to filter out a consistent boundary.
We then clip out the negative values of the result and refine it via the double-thresholding algorithm \cite{canny1986computational} to get the final contour.
As Laplacian operator is very sensitive to noise, bilateral filtering \cite{tomasi1998bilateral} is performed in advance to avoid the potential artifacts of input images.
%
Based on the extracted contour, we segment input frame into a bunch of color pieces. 
To realize this, we adopt the ``trapped-ball'' filling algorithm proposed in \cite{zhang2009vectorizing}, where each area enclosed by the contour is split as a single piece.
%
%

\noindent{\bf Network Structures of RFR Module} (Section 4.2).
In RFR, a feature extraction network (FeatureNet) transforms the input frames $I_0$ and $I_1$ into deep features $\mathcal F_0$ and $\mathcal F_1$, respectively, and a three-layer CNN is adopted to predict confidence maps to suppress the unreliable values in the coarse flows $f_{0\to1}$ and $f_{1\to0}$.
The detailed structures of the FeatureNet and the three-layer CNN are listed in Table \ref{structure}.

Then, a series of residues $\{\Delta f^{(t)}_{0\to1}\}$ are learnt via  a convolutional GRU \cite{convGRU}:
\begin{equation}
\begin{aligned}
x^{(t)} &= [f^{(t)}_{0\to1}, \mathcal F_0, corr(\mathcal F_0, \mathcal F^{(t)}_{1\to0})],\\
z^{(t)} &= \sigma(\text{Conv}([h^{(t-1)}, x^{(t)}])),\\
r^{(t)} &= \sigma(\text{Conv}([h^{(t-1)}, x^{(t)}])),\\
{\widetilde h}^{(t)} &=   \sigma(\text{Conv}([r^{(t)} \odot h^{(t-1)}, x^{(t)}])),\\
h^{(t)} &= (1 - z^{(t)}) \odot h^{(t - 1)} + z^{(t)} \odot {\widetilde h}^{(t)},\\
\Delta f^{(t)}_{0\to1} &= \text{Conv}(h^{(t)}),
\end{aligned}
\label{eq:bla}
\end{equation}
where Conv represents $3\times3$ convolutions, $\sigma$ denotes the ReLU.
For two $N\times H \times W$ features $\mathcal F_0$ and $\mathcal F_1$ ($N$, $H$, $W$ are the channel numbers, the height and the width of the tensors), the $corr$ operation computes a normalized correlation $c$ between those two tensors as 
\begin{equation}
c(u\cdot \Delta + v, i, j) = \sum_{n=0}^{N-1} \mathcal F_0(n, i, j) \cdot \mathcal F_1(n, i+u, j+v) /N,
\end{equation}
where $\Delta$ is the  shift size and $u, v \in [-\frac{\Delta-1}{2},+\frac{\Delta-1}{2} ]$.


\begin{table} [t]
    \centering
    \small{
    \caption{{\bf{Architectures of ``FeatureNet'' and ``3-layer CNN'' in RFR module.} ({Section 4.2}}). The arguments in Conv represent the input channel number, the output channel number, the kernel size, and the convolution stride in turn. RB denotes Residual Block where the input is downsampled to the same size of the learned residue. $\sigma$ represents ReLU. }\label{structure}
    \begin{tabular}{c|l}
     \bottomrule
     & \bf{Layer name(s)} \\
    \hline \hline
    ~ & Conv(3, 64, 7, 2), $\sigma$ \\
    ~ & RB[Conv(64, 64, 3, 1), $\sigma$, Conv(64, 64, 3, 1), $\sigma$] \\
    \multirow{2}*{\rotatebox{90}{FeatureNet}} & RB[Conv(64, 64, 3, 1), $\sigma$, Conv(64, 64, 3, 1), $\sigma$] \\
    &  RB[Conv(64, 96, 3, 2), $\sigma$, Conv(96, 96, 3, 1), $\sigma$] \\
    &  RB[Conv(96, 96, 3, 1), $\sigma$, Conv(96, 96, 3, 1), $\sigma$] \\
    &  RB[Conv(96, 128, 3, 2), $\sigma$, Conv(128, 128, 3, 1), $\sigma$] \\
    &  RB[Conv(128, 128, 3, 1), $\sigma$, Conv(128, 128, 3, 1), $\sigma$] \\
    &  Conv(128, 256, 1, 1)\\
    \hline
    \multirow{3}*{\rotatebox{90}{3-layer  $\,$}}  & Conv(6, 64, 5, 1), $\sigma$ \\
     & Conv(64, 32, 3, 1), $\sigma$ \\
    & Conv(32, 1, 3, 1), $\sigma$ \\
    \toprule
    \end{tabular}
    }

\end{table}

\begin{figure*}[t]
  \centering
    \small{
    \begin{tabular}{@{ }c@{ }c@{ }c@{ }c@{ }c@{ }c@{ }}
       
           \includegraphics[width=.15\linewidth]{figures/supp/cut_10_000662.jpg} \ &
        \includegraphics[width=.15\linewidth]{figures/supp/cut_10_000663.jpg}\ &
        \includegraphics[width=.15\linewidth]{figures/supp/cut_10_000662_super.jpg}\ &
        \includegraphics[width=.15\linewidth]{figures/supp/cut_10_000662_dain.jpg} \ &
        \includegraphics[width=.15\linewidth]{figures/supp/cut_10_000662_soft.jpg}\ &
        \includegraphics[width=.15\linewidth]{figures/supp/cut_10_000662_ours.jpg}\\
        \vspace{-12pt}\\

                \includegraphics[width=.15\linewidth]{figures/supp/cut_57_094090.jpg}\ &
        \includegraphics[width=.15\linewidth]{figures/supp/cut_57_094093.jpg}\ &
        \includegraphics[width=.15\linewidth]{figures/supp/cut_57_094090_3_super.jpg}\ &
        \includegraphics[width=.15\linewidth]{figures/supp/cut_57_094090_3_dain.jpg}\ &
        \includegraphics[width=.15\linewidth]{figures/supp/cut_57_094090_3_soft.jpg}\ &
        \includegraphics[width=.15\linewidth]{figures/supp/cut_57_094090_3_ours.jpg}\\
        \vspace{-12pt}\\

        \includegraphics[width=.15\linewidth]{figures/supp/cut_71_088679.jpg}\ &
        \includegraphics[width=.15\linewidth]{figures/supp/cut_71_088682.jpg}\ &
        \includegraphics[width=.15\linewidth]{figures/supp/cut_71_088679_3_super.jpg}\ &
        \includegraphics[width=.15\linewidth]{figures/supp/cut_71_088679_dain.jpg}\ &
        \includegraphics[width=.15\linewidth]{figures/supp/cut_71_088679_3_soft.jpg}\ &
        \includegraphics[width=.15\linewidth]{figures/supp/cut_71_088679_ours.jpg}\\
        \vspace{-12pt}\\
        \includegraphics[width=.15\linewidth]{figures/supp/cut_72_101836.jpg}\ &
        \includegraphics[width=.15\linewidth]{figures/supp/cut_72_101839.jpg}\ &
        \includegraphics[width=.15\linewidth]{figures/supp/cut_72_101836_2_super.jpg}\ &
        \includegraphics[width=.15\linewidth]{figures/supp/cut_72_101836_2_dain.jpg}\ &
        \includegraphics[width=.15\linewidth]{figures/supp/cut_72_101836_2_soft.jpg}\ &
        \includegraphics[width=.15\linewidth]{figures/supp/cut_72_101836_2_ours.jpg}\\
        \vspace{-12pt}\\

        Frame 0 & Frame 1 & Super SloMo & DAIN & SoftSplat   &  Ours  \\
    \end{tabular}
        
    \caption{\textbf{Visual comparisons with state-of-the-art methods.} Our proposed method generates more complete objects in the interpolation results. 
%
    }\label{fig:compare2}
    }

\end{figure*}

\section{More Experimental Results}
\noindent{\bf Performance on different motion tags} (Section 5.3).
In supplement to the experimental results of the main paper, the performance of various methods on different motion categories is shown in Figure \ref{fig:motion_tags}.
In general, the proposed AnimeInterp achieves the highest PSNR score on each motion category.
For specific motion categories, \eg, ``Walking'', AnimeInterp can improve over 0.4dB than state-of-the-art methods.
The influence of different motion categories on the performance of video interpolation can be further analyzed in the future studies.

\noindent{\bf More visual comparisons} (Section 5.1).
More visual comparisons between our proposed method and the state-of-the-art are shown in Figure \ref{fig:compare2}.
A video demo is also attached in the supplementary file to show the effectiveness of the proposed method.

{\small
\bibliographystyle{ieee_fullname}
\bibliography{egbib.bib}
}

%% file: content/intro.tex
%
%

In the animation industry, cartoon videos are produced from hand drawings of expert animators using a complex and precise procedure.
To draw each frame of an animation video manually would consume tremendous time, thus leading to a prohibitively high cost.
In practice, the animation producers usually replicate one drawing two or three times to reduce the cost, which results in the actual low frame rate of animation videos.
%
%
Therefore, it is highly desirable to develop computational algorithms to interpolate the intermediate animation frames automatically.

\begin{figure}[t]
\centering
 \includegraphics[width=0.9\linewidth]{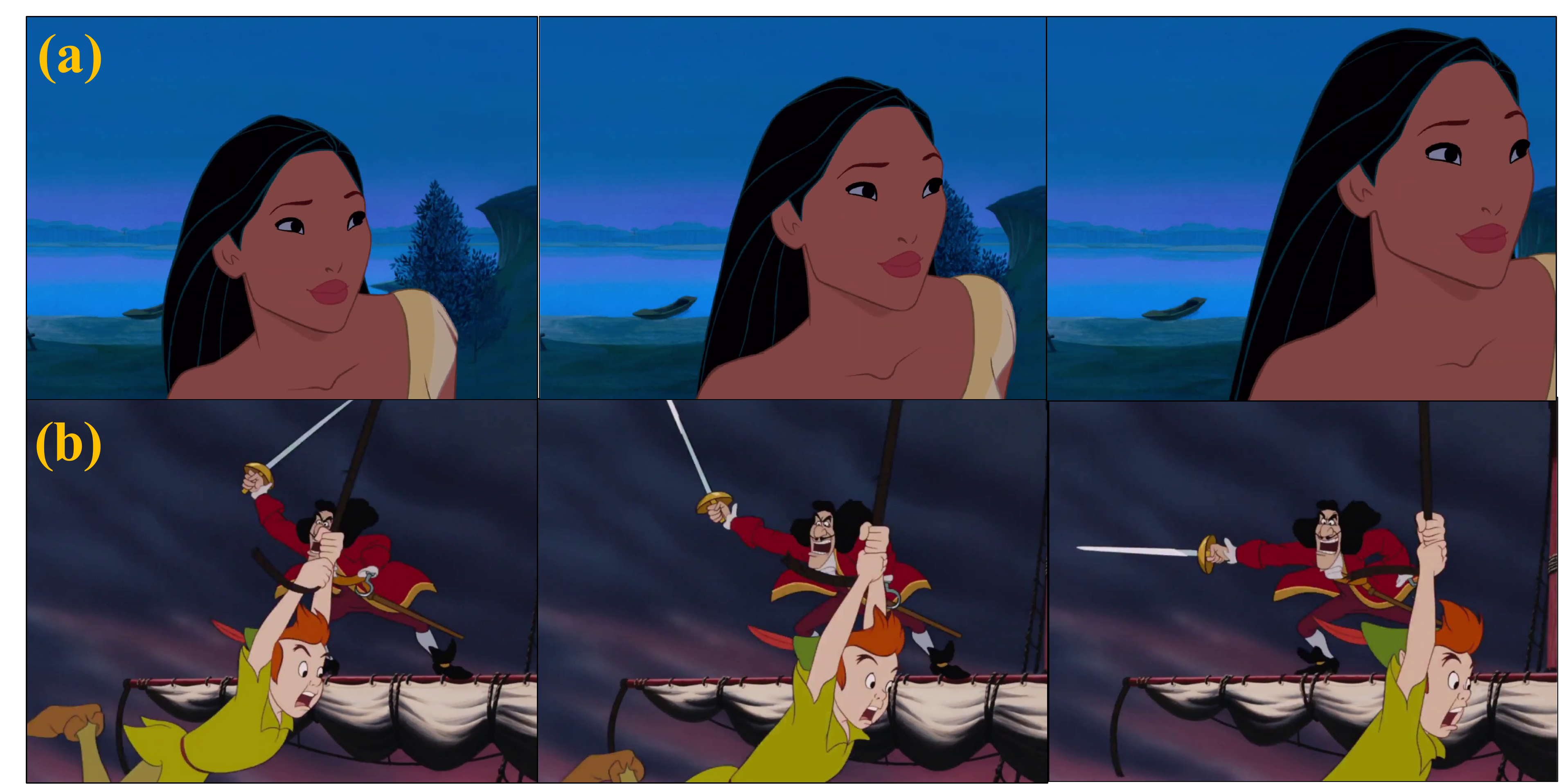}
\vspace{-6pt}
\caption{\textbf{Two challenges in animation video interpolation.} (a) Piece-wise smooth animations lack of textures. (b) Non-linear and extremely large motions.}\label{fig:difficulties}
\vspace{-15pt}
\end{figure}

In recent years, video interpolation has made great progress 
on natural videos.
%
%
However, in animations, existing video interpolation methods are not able to produce satisfying in-between frames.
An example from the film \textit{Children Who Chase Lost Voices} is illustrated in Figure~\ref{fig:teaser},
%
where the current state-of-the-art methods fail to generate a piece of complete luggage due to the incorrect motion estimation, which is shown in the lower-left corner of the image.
The challenges here stem from the two unique characteristics of animation videos:
%
%
1) First, cartoon images consist of explicit sketches and lines, which split the image into segments of smooth color pieces.
%
%
%
Pixels in one segment are similar, which yields insufficient textures to match the corresponding pixels between two frames and hence increases the difficulty to predict accurate motions. 
%
%
2) Second, cartoon animations use exaggerated expressions in pursuit of artistic effects, which result in non-linear and extremely large motions between adjacent frames.
Two typical cases are depicted in Figure \ref{fig:difficulties} (a) and (b) which illustrate these challenges respectively.
%
%
Due to these difficulties mentioned above, video interpolation in animations remains a challenging task.

In this work, we develop an effective and principled novel method for video interpolation in animations.
%
In particular, we propose an effective framework, \textbf{AnimeInterp}, to address the two aforementioned challenges.
AnimeInterp consists of two dedicated modules: a Segment-Guided Matching (SGM) module and a Recurrent Flow Refinement (RFR) module, which are designed to predict accurate motions for animations in a coarse-to-fine manner.
%
%
More specifically, the proposed SGM module computes a coarse piece-wise optical flow using global semantic matching among color pieces split by contours.
%
%
Since the similar pixels belonging to one segment are treated as a whole, SGM can avoid the local minimum caused by mismatching on smooth areas, which resolves the ``lack of textures'' problem.
To tackle the ``non-linear and extremely large motion'' challenge in animation, the piece-wise flow estimated by SGM is further enhanced by a Transformer-like network named Recurrent Flow Refinement.
As shown in Figure~\ref{fig:teaser}, our proposed approach can better estimate the flow of the luggage in large displacement and generate a complete in-between frame.



%
A large-scale animation triplet dataset, \textbf{ATD-12K}, is built to facilitate comprehensive training and evaluations of video interpolation methods on cartoon videos.
Unlike other animation datasets, which consists of only single images, ATD-12K contains 12,000 frame triplets selected from 30 animation movies in different styles with a total length over 25 hours.
%
%
%
Apart from the diversity, our test set is divided into three difficulty levels 
according to the magnitude of the motion and occlusion. 
We also provide annotations on movement categories for further analysis.
%


The contributions of this work can be summarized as follows:
\textbf{1)} We formally define and study the animation video interpolation problem for the first time. This problem has significance to both academia and industry.
\textbf{2)} We propose an effective animation interpolation framework named AnimeInterp with two dedicated modules to resolve the ``lack of textures'' and ``non-linear and extremely large motion'' challenges. Extensive experiments demonstrate that AnimeInterp outperforms existing state-of-the-art methods both quantitatively and qualitatively.
\textbf{3)} We build a large-scale cartoon triplet dataset called ATD-12K with large content diversity representing many types of animations to test animation video interpolation methods. Given its data size and rich annotations, ATD-12K would pave the way for future research in animation study.




%% file: content/related.tex
%
%

\noindent{\bf Video Interpolation.} 
%
%
%
%
Video interpolation has been widely studied in recent years.
In \cite{meyer2015phase}, Meyer \etal  propose a phase-based video interpolation scheme, which performs impressively on videos with small displacements.
In  \cite{niklaus2017video,niklaus2017video2}, Niklaus \etal design a kernel-based framework to sample interpolated pixels via convolutions on corresponding patches of adjacent frames.
%
%
%
However, the kernel-based framework is still not capable of processing large movements due to the limitation of the kernel size. 
To tackle the large motions in videos, many studies use optical flows for video interpolation.
%
%
Liu \etal \cite{liu2017video} predict a 3D voxel flow  to sample the inputs into the middle frame.
Similarly, Jiang \etal \cite{jiang2018super} propose to jointly estimate bidirectional flows and an occlusion mask for multi-frame interpolation.
Besides, some studies are dedicated to improving warping and synthesis with given bidirectional flows \cite{bao2018memc, bao2019depth, niklaus2018context, niklaus2020softmax}, and higher-order motion information is used to approximate real-world videos \cite{xu2019quadratic, chi2020all}. 
%
%
%
%
In addition to using  pixel-wise flows on images, ``feature flows'' on deep features are also explored for video interpolation \cite{gui2020featureflow}.
%
Although existing methods achieve great success in interpolating real-world videos , they fail to handle the large and non-linear motions of animations. Thus, the animation video interpolation still remains unsolved.
%
%
%
%
In this paper, we propose a segment-guided matching module based on the color piece matching, boosting the flow prediction. 

\noindent{\bf Vision for Animation.}
%
In vision and graphics community, there are many works on processing and enhancing the animation contents, \eg, manga sketch simplification \cite{simo2016learning,simo2018mastering}, vectorization of cartoon images \cite{zhang2009vectorizing,yao2016manga}, stereoscopy of animation videos \cite{liu2013stereoscopizing}, and colorization of black-and-white manga \cite{qu2006manga,sykora2004unsupervised,kopf2012digital}.
In recent years, with the surge of deep learning, researchers attempt to generate favorable animation contents.
For example, generative adversarial networks are used to generate vivid cartoon characters \cite{jin2017towards,xiang2019disentangling} and style-transfer models are developed to transfer real-world images to cartoons \cite{chen2018cartoongan,wang2020learning,chen2019animegan}.
However, generating animation video contents is still a challenging task due to the difficulty of estimating the motion between frames.

%% file: content/dataset.tex
To facilitate the training and evaluation of animation video interpolation methods, we build a large-scale dataset named ATD-12K, which comprises a training set containing 10,000 animation frame triplets and a test set containing 2,000 triplets, collected from a variety of cartoon movies.
%
To keep the objectivity and fairness, test data are sampled from different sources that are exclusive from the training set.
Besides, to evaluate video interpolation methods from multiple aspects, we  provide rich annotations to the test set, including levels of difficulty, the Regions of Interest (RoIs) on movements, and tags on motion categories. 
Some typical examples of ATD-12K and the annotation information are displayed in Figure~\ref{fig:ani_triplets}.


    %
    

\subsection{Dataset Construction}
\vspace{-1pt}
In the first step to build ATD-12K, we download a large number of animation videos from the Internet.
To keep the diversity and representativeness of ATD-12K, a total of 101 cartoon clips are collected from 30 series of movies made by diversified producers, including Disney, Hayao, Makoto, Hosoda, Kyoto Animation, and A1 pictures, with a total duration of more than 25 hours.
The collected videos have a high visual resolution of $1920\times1080$ or $1280\times720$.
%
%
Second, we extract sufficient triplets from the collected videos.
%
In this step, we not only sample adjacent consecutive frames into the triplet but also extract those with one or two frames as an  interval to enlarge the inter-frame displacement.
%
%
Since the drawing of an animation frame may be duplicated several times, it is common to sample similar contents in the triplets extracted from one video.
%
%
To avoid high affinity among our data, we filter out the triplets that contains two frames with a structural similarity (SSIM) \cite{wang2004image} larger than 0.95.
Meanwhile, triplets with SSIM lower than 0.75 are also dropped off to get rid of scene transitions in our data.
%
After that, we manually inspect the remaining 131,606 triplets for further quality improvement.
%
%
Triplets that are marked as eligible by at least two individuals are kept.
Frames that contain inconsistent captions, simple or similar scenes or untypical movements are removed.
After the strict selection, 12,000 representative animation triplets are used to construct our final ATD-12K.

\noindent{\bf Statistics.} 
To explore the gap between natural and cartoon videos, we compare the motion statistics of the ATD-12K and a real-world dataset Adobe240 \cite{su2017deep}, which is often used as an evaluation set for video interpolation. 
%
We estimate the optical flow between frames of every input pair in the two datasets, and compute the mean and standard deviation of displacements for each flow.
The normalized histograms of means and the standard deviations in the two datasets are shown in Figure~\ref{fig:ani_triplets}.
They reveal that the cartoon dataset ATD-12K has a higher percentage of large and diverse movements than the real-world Adobe240 dataset.

\vspace{-1pt}
\subsection{Annotation}\label{subsect:annotaion}
\vspace{-1pt}


\noindent{\bf Difficulty Levels.} 
%
 We divided the test set of ATD-12K into three difficulty levels, \ie, ``Easy'', ``Medium'', and ``Hard'' based on the average magnitude of motions and the area of occlusion in each triplet.
 For more details on the definition of each level, please refer to the supplementary file. Sample triplets of different levels are displayed in Figure~\ref{fig:ani_triplets}.


\begin{figure*}[h]
    \scriptsize
    \setlength{\tabcolsep}{1.5pt}
    \centering
    \includegraphics[width=0.95\linewidth]{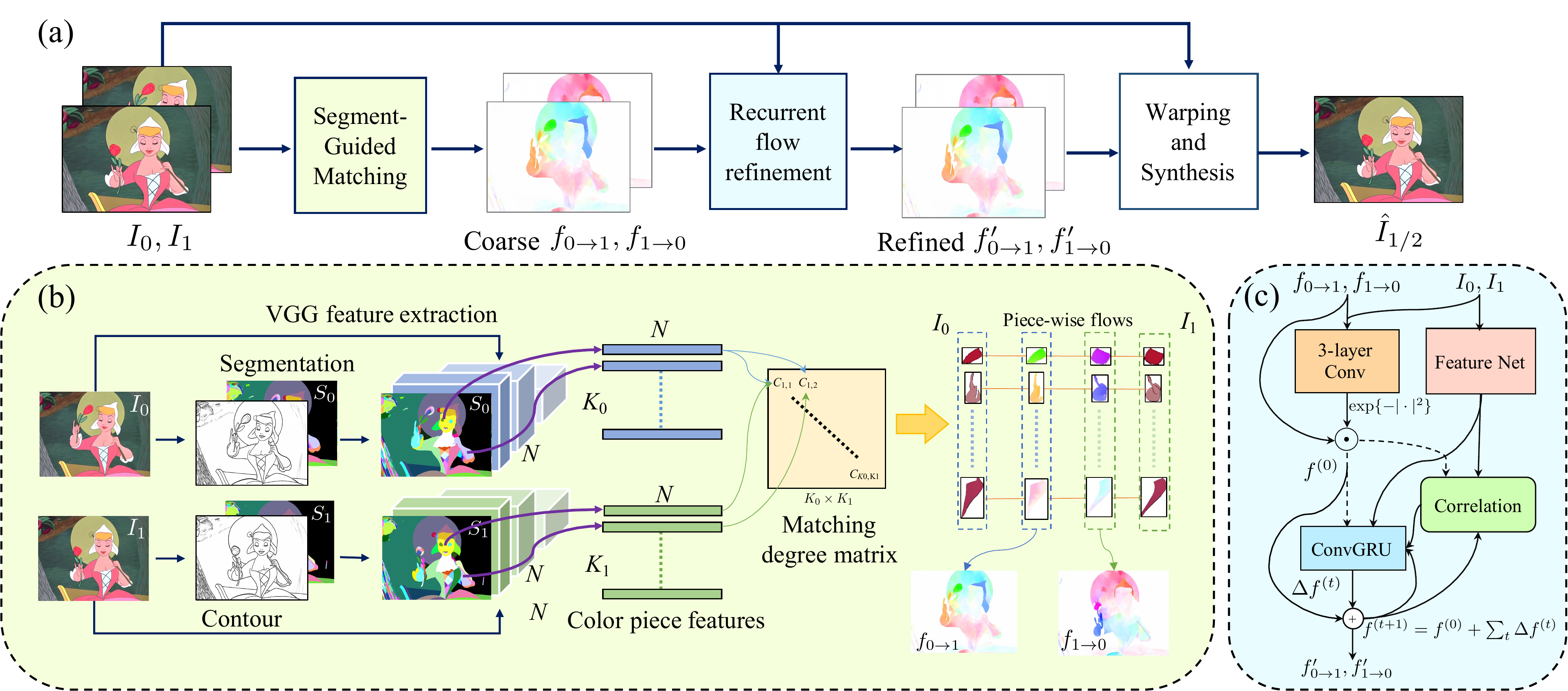}\\
    \vspace{-3pt}
    \caption{\textbf{(a) The overall pipeline of AnimeInterp.} The input $I_0$ and $I_1$ are fed to the SGM module to generate the coarse flows, \ie $f_{0\to1}$ and $f_{1\to0}$. Then, $f_{0\to1}$ and $f_{1\to0}$ are refined by the RFR module. The final interpolation result is produced by the warping and synthesis network borrowed from SoftSplat \cite{niklaus2020softmax}. \textbf{(b) The inner structure of the SGM module.} \textbf{(c) The workflow of the RFR module.}}
    \label{fig:pipeline}
    \vspace{-14pt}
\end{figure*}

\noindent{\bf Motion RoIs.} 
Animation videos are made of moving foreground objects and motion-less background scenes. 
Since the foreground characters are more visually appealing, the motions of those regions have more impact on audiences' visual experience.
To better reveal the performance of interpolation methods on motion areas, we provide a bounding box for the 2$^{nd}$ image (as shown in Figure~\ref{fig:ani_triplets} test set) of each triplet to delineate the motion RoI. 

\noindent{\bf Motion Tags.} 
We also provide tags to describe the major motion of a triplet. Our motion tags can be classified into two categories, namely, 1) general motion types including translation, rotation, scaling, and deformation; 2) character behaviors containing speaking, walking, eating, sporting, fetching, and others.
The percentage for tags in each category is shown in Figure~\ref{fig:ani_triplets}.

%% file: content/algorithm.tex

\vspace{-3pt}
The overview of our framework is shown in Figure \ref{fig:pipeline}. 
%
Given the input images $I_0$ and $I_1$, we first estimate the coarse flows $f_{0\to1}$ and $f_{1\to0}$ between $I_0$ and $I_1$ in both directions via the proposed SGM module in Section~\ref{subsect:guidance-flow}.  
%
%
Then, we set the coarse flow as the initialization of a recurrent neural network and gradually refine it to obtain the fine flows $f_{0\to1}^{'}$ and $f_{1\to0}^{'}$ in Section~\ref{subsect:rrnn-flow}.
Finally, we warp $I_0$ and $I_1$ using $f_{0\to1}^{'}$ and $f_{1\to0}^{'}$ and synthesize the final output $\hat{I}_{1/2}$ in Section~\ref{subsect:synthesis}.
%

    
\vspace{-2pt}
\subsection{Segment-Guided Matching}
\label{subsect:guidance-flow}
For typical 2D animations, objects are usually outlined with explicit lines, and each enclosed area is filled with a single color.
The color of the moving object in one frame remains stable in the next frame despite the large displacement, which could be regarded as a strong clue to find appropriate semantic matching for the color pieces.
In this paper, we leverage this clue to tackle smooth piece-wise motions by generating coarse optical flows. The procedure is illustrated in Figure~\ref{fig:pipeline}(b), and we elaborate it in the following. 
%

\noindent{\bf Color Piece Segmentation.}
Following Zhang \etal's work \cite{zhang2009vectorizing}, we adopt the Laplacian filter to extract contours of animation frames. Then, the contours are filled by the ``trapped-ball'' algorithm \cite{zhang2009vectorizing} to generate color pieces.
After this step, we obtain a segmentation map $S\in \mathbb N^{H\times W}$, where pixels of each color piece is labeled by an identity number.
In the rest of this section, we notate the segmentation of the input $I_0$ and $I_1$ as $S_0$ and $S_1$, respectively. $S_0(i)$ represents the pixels in the $i^{th}$ color piece of $I_0$, and $S_1(i)$ for $I_1$ is similar.


\noindent{\bf Feature Collection.}
We feed the input $I_0$ and $I_1$ into a pretrained VGG-19 model \cite{VGG} and extract features of  \texttt{relu1\_2},  \texttt{relu2\_2}, \texttt{relu3\_4} and \texttt{relu4\_4} layers.
Then, we assemble the features belonging to one segment by the super-pixel pooling proposed in \cite{liu2015learning}.
%
Features of smaller scales are pooled by the down-sampled segmentation maps and are concatenated together.
After the pooling, each color piece is represented by an $N$-dimensional feature vector, and the whole image is reflected to a $K\times N$ feature matrix, where $K$ is the number of the color pieces and each row of the matrix represents the feature of the corresponding color piece. The feature matrices $I_0$ and $I_1$ are denoted as $F_0$  and  $F_1$, respectively.
%


\noindent{\bf Color Piece Matching.}
%
%
We now use $F_0$ and $F_1$ to estimate a consistent mapping between color pieces of $I_0$ and $I_1$.
Specifically, we predict a forward map $\mathcal M_{0\to1}$ and a backward map $\mathcal M_{1\to0}$.  
For the $i^{th}$ color piece in the first frame, the forward map $\mathcal M_{0\to1}(i)$ indicates the maximum-likelihood corresponding piece in the second frame and the backward map does the same from $I_1$ to $I_0$.
%
%
%
%
To quantify the likelihood of a candidate pair $(i\in S_0, j \in S_1)$, we compute an affinity metric $\mathcal A$ using $F_0$ and $F_1$ as
\vspace{-5pt}
\begin{equation}
\label{eq:correspondance}
\mathcal{A}(i, j)= \sum_n^{N}{\rm{min}}\left(\tilde F_0(i, n), \tilde F_1(j, n)\right),
\end{equation}
where $\tilde F_0(i, n) = F_0(i, n) / \sum_n F_0(i, n)$ is the normalized feature of $F_0$, and $\tilde F_1$ is similar to $\tilde F_0$.
This affinity metric measures the similarities of all pairs globally.
%
%
Besides, 
to avoid potential outliers,
we also exploit two constraint penalties, \ie, the {\it{distance penalty}} and the {\it{size penalty}}.
%
First, we assume that the displacement between two corresponding pieces is unlikely to be overly large.
The {\it{distance penalty}} is defined as the ratio of the distance between the centroids of two color pieces and the diagonal length of the image, 
%
\begin{equation}
\label{eq:disdance_loss}
{\mathcal L_{dist}}(i, j)=  \frac{\|P_0(i) - P_1(j)\| }{\sqrt{H^2 + W^2}}, 
\end{equation}
where $P_0(i)$ and  $P_1(j)$ represent the positions of the centroids of $S_0(i)$ and $S_1(j)$, respectively.
Note that this penalty is only performed to the matching with the displacement larger than $15\%$ of the diagonal length of the image.
Second, we suggest that the sizes of matched pieces should be similar.
The {\it{size penalty} }is designed as, 
\begin{equation}
\label{eq:correspondance}
{\mathcal L_{size}}(i, j) =  \left|\frac{|S_0(i)| - |S_1(j)|}{HW}\right|,
\end{equation}
%
where $|\cdot|$ denotes the number of pixels in a cluster.

Combining all items above, a matching degree matrix $\mathcal C$ is then computed as, 
\begin{equation}
\label{eq:c+d+n}
 \mathcal C = \mathcal A - \lambda_{dist}\mathcal L_{dist} - \lambda_{size}{\mathcal L_{size}}, 
\end{equation}
where $\lambda_{dist}$ and $\lambda_{size}$ are set to $0.2$ and $0.05$ in our implementation, respectively.
For each pair $(i\in S_0, j\in S_1)$, $\mathcal C(i, j)$ indicates the likelihood. 
Hence, for the $i^{th}$ color piece in $S_0$, the most likely matching piece in $S_1$ is the one with the maximum matching degree, and vice versa. The mapping of this matching is formulated as, 
\begin{equation}
\begin{aligned}
 \mathcal M_{0\to 1}(i) = \arg\max_j(\mathcal C (i, j)),\\
 \mathcal M_{1\to 0}(j) = \arg\max_i(\mathcal C (i, j)).
\end{aligned}
\label{eq:m1}
\end{equation}
%
%

\noindent{\bf Flow Generation.}
%
In the last step of SGM module, we predict dense bidirectional optical flows $f_{0\to1}$ and $f_{1\to0}$ using $\mathcal M_{0\to 1}$ and $\mathcal M_{1\to 0}$.
We only describe the procedure to compute $f_{0\to1}$, for $f_{1\to0}$ can be obtained by reversing the mapping order.
For each matched color-piece pair $(i, j)$ where $j=\mathcal M_{0\to1}(i)$, we first compute the displacement of the centroids as a shift base $f_c^i=P_1(j) - P_0(i)$, and then we compute the local deformation $f^i_d:=(u,v)$ by variational optimization, 
\vspace{-1pt}
\begin{equation}
\begin{aligned}
    E\left(f_d^i(\mathbf x)\right)= &\int{|I^j_1\left(\mathbf x + f_{c}^i(\bm x) + f_d^i(\mathbf x)\right) - I_0^i(\mathbf x)|d\mathbf x} \\ 
    & + \int{\left(|\nabla u(\mathbf x)|^2 + \nabla |v(\mathbf x)|^2 \right)|d\mathbf x}.
\end{aligned}
\end{equation}
Here, $I^i_0$ represents a masked $I_0$ where pixels not belonging to $i^{th}$ color piece are set to $0$. $I^j_1$ is the similar to $I^i_0$.
The optical flow for pixels in $i^{th}$ color piece is then $f_{0\to1}^i = f_d^i + f_c^i$.
%
%
%
Since the color pieces are disjoint, the final flow for the whole image is calculated by simply adding all piece-wise flows together as  $f_{0\to1}=\sum_i f_{0\to1}^i$.

To further avoid outliers, we mask the flow of the $i^{th}$ piece to zero if it does not satisfy the mutual consistency, \ie,
$ 
\mathcal M_{1\to0}( \mathcal M_{0\to1}(i)) \neq i
$
.
%
This operation prevents the subsequent flow refinement step to be misled by the low-confidence matching.

\begin{table*}
    
\caption{\textbf{Quantitative results on the test set of ATD-12K.} The best and runner-up values are bold and underlined, respectively.}
\vspace{-8pt}
  
  \centering
  
\small{
  \begin{tabular}{l  c c c c c c c c c c }
    \toprule
    &  \multicolumn{2}{c}{ Whole } & \multicolumn{2}{c}{ RoI } & \multicolumn{2}{c}{ Easy } & \multicolumn{2}{c}{ Medium } & \multicolumn{2}{c}{ Hard }             \\
     \cmidrule(r){2-3} \cmidrule(r){4-5} \cmidrule(r){6-7} \cmidrule(r){8-9} \cmidrule(r){10-11} 
    \small Method &PSNR &SSIM &  PSNR &   SSIM &    PSNR &   SSIM &     PSNR &   SSIM &   PSNR & SSIM \\
      \midrule
  \rowcolor{mygray}
  Super SloMo w/o. ft.    &  27.88 & 0.946 &  24.56 & 0.886 & 30.66 & 0.966 & 27.29 & 0.948 & 24.63 & 0.917             \\
  Super SloMo \cite{jiang2018super}   & 28.19 & 0.949 & 24.83 & 0.892 & 30.86 & 0.967 & 27.63 & 0.950 & 25.02 &  0.922            \\
  \rowcolor{mygray}
  DAIN w/o. ft.      & 28.84 & 0.953 & 25.43 & 0.897 & 31.40 & \underline{0.970} & 28.38 & 0.955 & 25.77 & 0.927  \\ 
  DAIN \cite{bao2019depth}  & 29.19 & 0.956 & 25.78 & 0.902 & 31.67 & \textbf{0.971} & 28.74 & 0.957 & 26.22 & 0.932\\ 
  \rowcolor{mygray}
  QVI w/o. ft.  & 28.80 & 0.953 & 25.54 & 0.900 & 31.14 & 0.969 & 28.44 & 0.955 & 25.93 & 0.929     \\
  QVI \cite{xu2019quadratic} & 29.04 & 0.955 & 25.65 & 0.901 & 31.46 & \underline{0.970} & 28.63 & 0.956 & 26.11 & 0.931       \\

\rowcolor{mygray}
  AdaCoF w/o. ft.  & 28.10 & 0.947 & 24.72 & 0.886 & 31.09 & 0.968 & 27.43 & 0.948 & 24.65  & 0.916       \\
  AdaCoF \cite{lee2020adacof} & 28.29 & 0.951 & 24.89 & 0.894 & 31.10 & 0.969 & 27.63 & 0.951 & 25.10 & 0.925      \\
  \rowcolor{mygray}
  SoftSplat  w/o. ft.  & 29.15 & 0.955 & 25.75 & 0.904 & 31.50 & \underline{0.970} & 28.75 & 0.956 & 26.29 & 0.934    \\
  SoftSplat \cite{niklaus2020softmax} & 29.34 & \underline{0.957} & 25.95 & \underline{0.907} & 31.60 & \underline{0.970} & 28.96 & \underline{0.958} & 26.59 & \underline{0.938}    \\
  
  \midrule
  Ours w/o. SGM   & \underline{29.54} & {\bf 0.958} & \underline{26.15} &  {\bf 0.910} & \underline{31.80} & {\bf 0.971} & \underline{29.15} & {\bf 0.959}  & \underline{26.78} & {\bf{0.939} }      \\
    Ours w/o. RFR   & 27.62 & 0.944 &24.43 & 0.887 & 29.78 & 0.959 & 27.29 & 0.946 & 24.94 & 0.920         \\
  
  Ours   & {\bf 29.68} &  {\bf 0.958} & {\bf 26.27} & {\bf 0.910} & {\bf 31.86} & {\bf 0.971}  & {\bf 29.26} & {\bf 0.959} & {\bf 27.07} & {\bf 0.939}  \\

    \bottomrule
  \end{tabular}
  }
  \label{table:quantitative_eval}
   \vspace{-8pt}
\end{table*}

\subsection{Recurrent Flow Refinement Network}
\label{subsect:rrnn-flow}
%
In this section, we refine the coarse optical flows $f_{0\to1}$ and $f_{1\to0}$ to the finer views $f'_{0\to1}$ and $f'_{1\to0}$ via a deep Recurrent Flow Refinement (RFR) network.
%
There are two main motivations for introducing the RFR module.
%
First, since a strict mutual consistency constraint is adopted in the color-piece-matching step, non-robust pairs are masked off, leaving null flow values in some locations. RFR is able to produce valid flows for those locations.
%
%
Second, the SGM module is beneficial for large displacements but is less effective to predict precise deformation for the non-linear and exaggerated motion in animation videos.
Here, RFR complements the previous step by refining the coarse and piece-wise flows.

%
Inspired by the state-of-the-art optical flow network RAFT \cite{teed2020raft}, we design a transformer-like architecture as shown in Figure \ref{fig:pipeline}(c) to recurrently refine the piece-wise flow $f_{0\to1}$.
For the sake of brevity, we only illustrate the process to compute $f'_{0\to1}$. 
First,  to further avoid potential errors in the coarse flow $f_{0\to1}$,  a pixel-wise confidence map $g$ is learned to mask the unreliable values via a three-layer CNN, which takes concatenated $|I_1(\mathbf x + f_{0\to1}(\mathbf x )) - I_0(\mathbf x) |$, $I_0$ and $f_{0\to 1}$ as input.
Then, the coarse flow $f_{0\to 1}$ is multiplied with $\exp\{-g^2\}$ to be the initialization $f^{(0)}_{0\to1}$ for the following refinement.
Next, a series of residues $\{\Delta f^{(t)}_{0\to1}\}$ are learnt via  a convolutional GRU \cite{convGRU}:
\begin{equation}
    \Delta f^{(t)}_{0\to1} = \text{ConvGRU}\left(f^{(t)}_{0\to1}, \mathcal F_0, corr(\mathcal F_0, \mathcal F^{(t)}_{1\to0})\right),
\end{equation}
where $corr(\cdot,\cdot)$ calculates the correlation between two tensors, $\mathcal F_0$ and $\mathcal F_1$ are frame features extracted by a CNN,  and $\mathcal F^{(t)}_{1\to0}$ is bilinearly sampled from $\mathcal F_1$ with optical flow $f^{(t)}_{0\to1}$. 
The learnt residues are recurrently accumulated to update the initialization.
The optical flow refined after $T$ iterations is computed as
%
\vspace{-2pt}
\begin{equation}
   f^{(T)}_{0\to1} = f^{(0)}_{0\to1} + \sum_{t=0}^{T-1}\Delta f^{(t)}_{0\to1} .
\end{equation}
And the fine flow $f'_{0\to1}$ is the output of the last iteration.

\subsection{Frame Warping and Synthesis}
\label{subsect:synthesis}
To generate the intermediate frame using flow $f'_{0\to1}$ and $f'_{1\to0}$, we adopt the splatting and synthesis strategy of SoftSplat \cite{niklaus2020softmax}. 
%
%
In short, a bunch of features are extracted from $I_0$ and $I_1$ using a multi-scale CNN, and then, all features and input frames are splatted  via forward warping to the center position, \eg, $I_0$ is splatted to  $I_{0\to1/2}$ as
\begin{equation}
    I_{0\to 1/2} (\mathbf x + f_{0\to1}(\mathbf x )/2 ) = I_{0}(\mathbf x ).
\end{equation}
%
%
%
%
Finally, all warped frames and features are fed into a GridNet \cite{fourure2017residual} with three scale levels to synthesize the target frame $\hat I_{{1}/{2}}$. 
%

\subsection{Learning}
\label{sec:train}

To supervise the proposed network, we adopt the $L_1$ distance $\|\hat I_{1/2} - I_{1/2}\|_1$ between the prediction $\hat I_{1/2}$ and the ground truth $I_{1/2}$ as the training loss.

We train this network in two phases: the {\it training phase} and the {\it fine-tuning phase}.
In the {\it training phase}, we first pre-train the recurrent flow refinement (RFR) network following \cite{teed2020raft}, and then fix the weights of RFR to train the rest parts of proposed network on a real-world dataset proposed in \cite{xu2019quadratic} for 200 epochs. 
During this phase, we do not use the coarse flows predicted by SGM module.
The learning rate is initialized as $10^{-4}$ and decreases with a factor $0.1$ at the 100$^{th}$ and the $150^{th}$ epochs.
In the second phase, we fine-tune the whole system for another 50 epochs on the training set of our ATD-12K with the learning rate of $10^{-6}$.
During fine-tuning, images are rescaled into $960\times540$ and are randomly cropped into $380\times380$ with batch size 16.
%
We also stochastically flip the images and reverse the triplet order as data augmentation.



%

%% file: content/experiment.tex

\noindent\textbf{Compared Methods.}
AnimeInterp is compared with five most recent state-of-the-art methods including Super SloMo \cite{jiang2018super}, DAIN \cite{bao2019depth}, QVI \cite{xu2019quadratic}, AdaCoF \cite{lee2020adacof} and SoftSplat \cite{niklaus2020softmax}.
%
%
We use the original implementation of DAIN, QVI and AdaCoF, and the implementation from \cite{superslomo-reim} for Super SloMo.
Since no model is released for SoftSplat, we implement this method following the description in \cite{niklaus2020softmax}, and train it with the same strategy as our proposed model.
%
%
%
%
%

\noindent\textbf{Datasets.}
We fine-tune these baseline models on the training set of ATD-12K with the hyper-parameters described in Section \ref{sec:train}.
Models before and after fine-tuning are separately evaluated on our proposed ATD-12K test set. 

\noindent\textbf{Evaluation Metrics.}
We use two distant images in every test triplet as input frames and use the middle one as ground truth.
The PSNR and SSIM \cite{ssim} between the predicted intermediate results and the ground truth are adopted as evaluation metrics.
%
%
Using the annotation in Section \ref{subsect:annotaion}, we evaluate the interpolation results not only on the whole image (denoted as ``Whole''), but also on the regions of interest (denoted as ``RoI''). 
Meanwhile, we show the evaluations for the three levels (``Easy'', ``Medium'' and ``Hard'').

\begin{figure*}[t]
  \centering
    \small{
    \begin{tabular}{@{ }c@{ }c@{ }c@{ }c@{ }c@{ }c@{ }}
       
           \includegraphics[width=.142\linewidth]{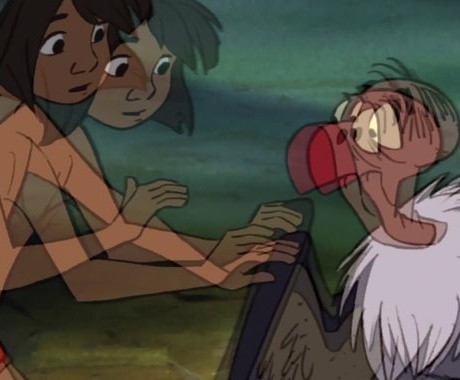}\ &
        \includegraphics[width=.142\linewidth]{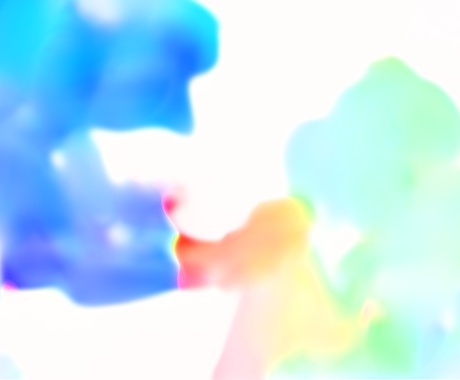}\ &
        \includegraphics[width=.142\linewidth]{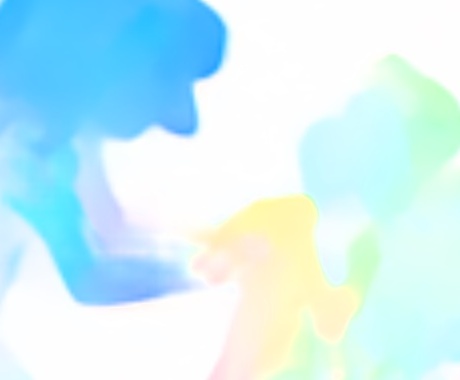}\ &
        \includegraphics[width=.142\linewidth]{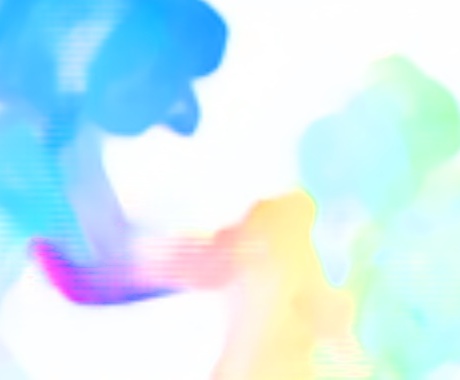}\ &
        \includegraphics[width=.142\linewidth]{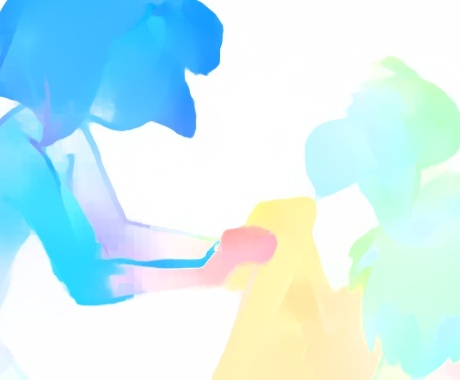}\ &
        \includegraphics[width=.142\linewidth]{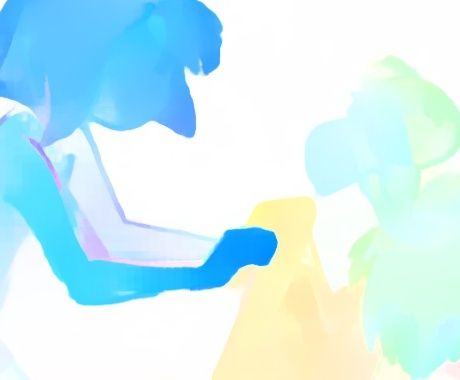}\\
        \vspace{-13pt}\\

                \includegraphics[width=.142\linewidth]{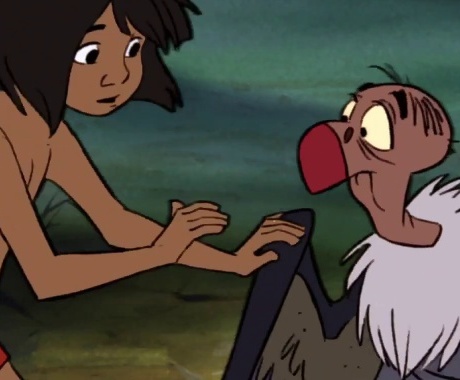}\ &
        \includegraphics[width=.142\linewidth]{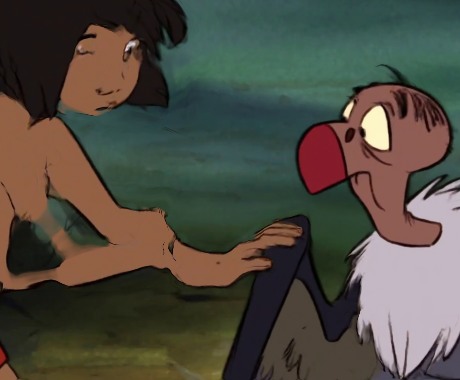}\ &
        \includegraphics[width=.142\linewidth]{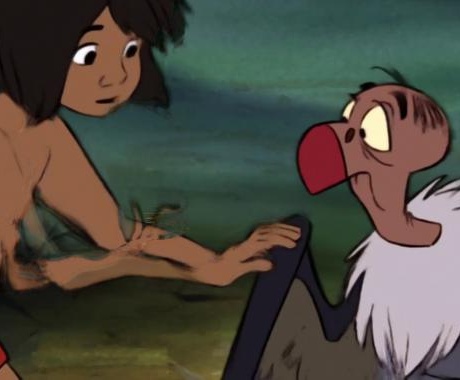}\ &
        \includegraphics[width=.142\linewidth]{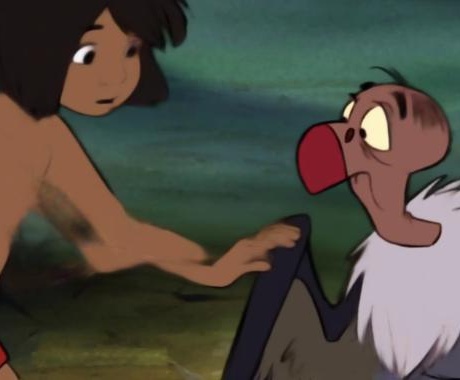}\ &
        \includegraphics[width=.142\linewidth]{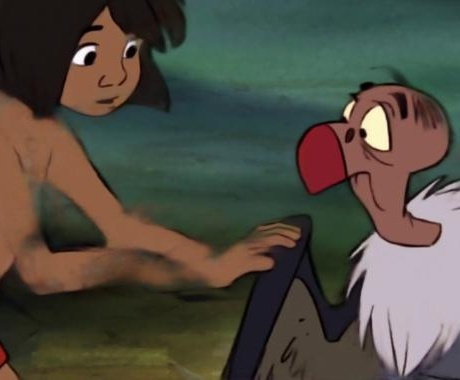}\ &
        \includegraphics[width=.142\linewidth]{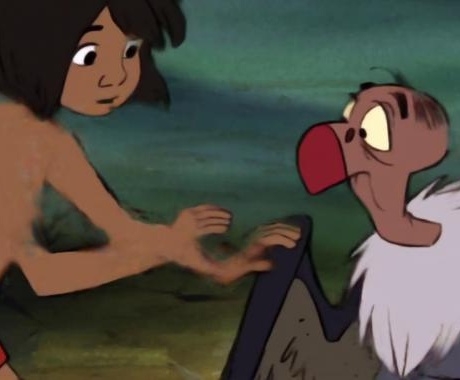}\\
        \vspace{-12pt}\\

        \includegraphics[width=.142\linewidth]{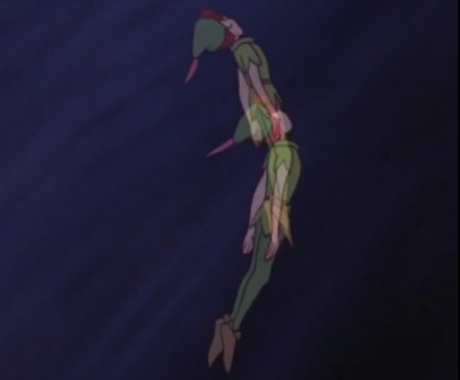}\ &
        \includegraphics[width=.142\linewidth]{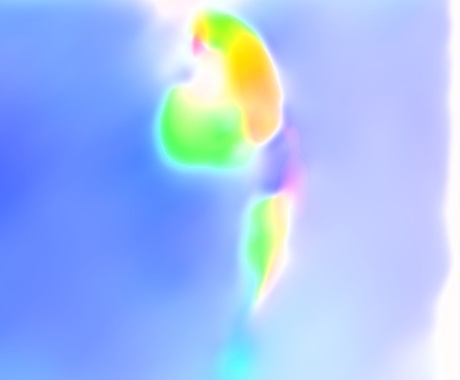}\ &
        \includegraphics[width=.142\linewidth]{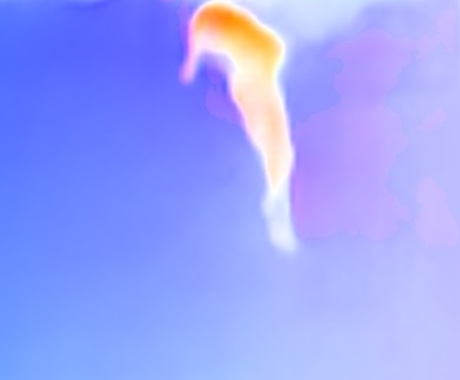}\ &
        \includegraphics[width=.142\linewidth]{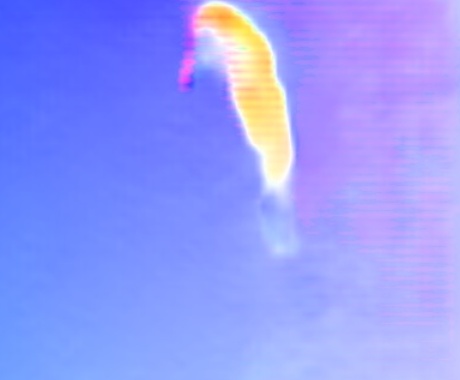}\ &
        \includegraphics[width=.142\linewidth]{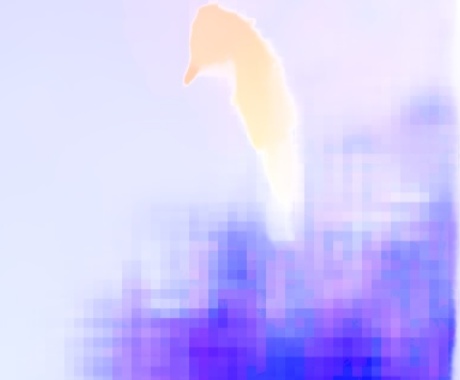}\ &
        \includegraphics[width=.142\linewidth]{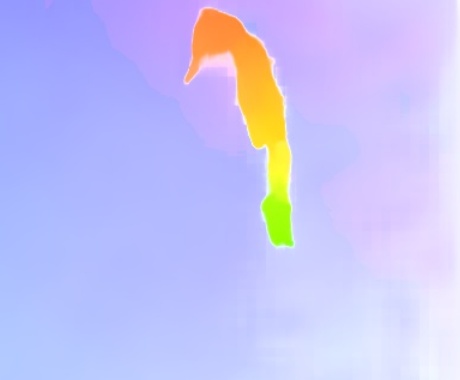}\\
        \vspace{-13pt}\\
        \includegraphics[width=.142\linewidth]{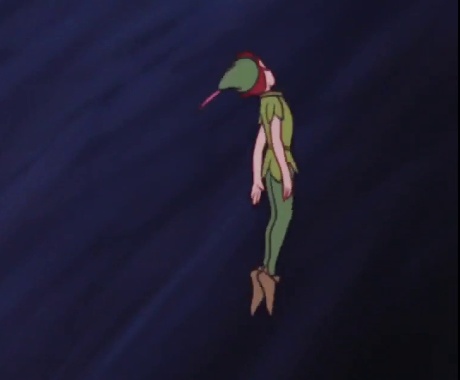}\ &
        \includegraphics[width=.142\linewidth]{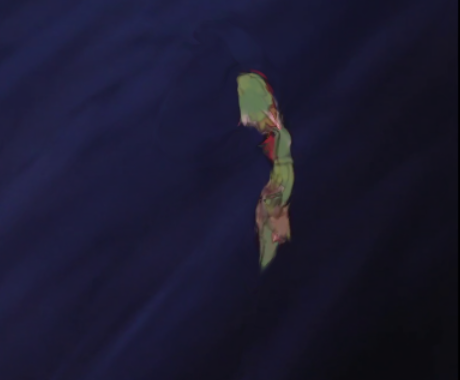}\ &
        \includegraphics[width=.142\linewidth]{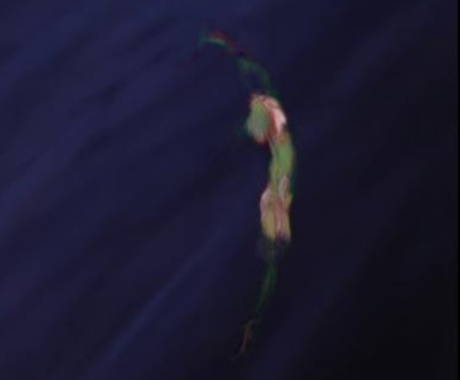}\ &
        \includegraphics[width=.142\linewidth]{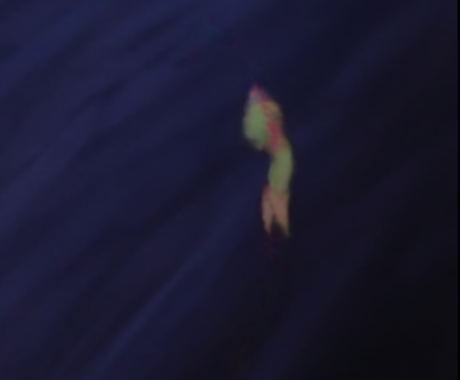}\ &
        \includegraphics[width=.142\linewidth]{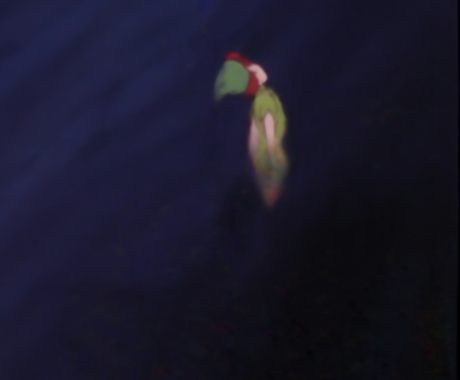}\ &
        \includegraphics[width=.142\linewidth]{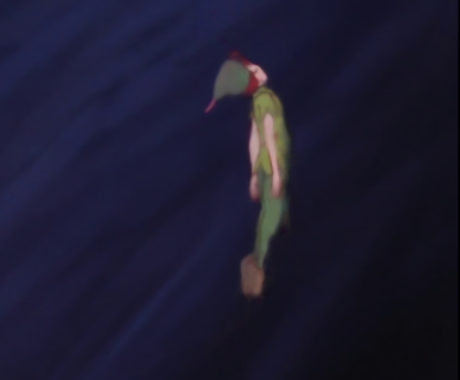}\\
        \vspace{-12pt}\\

        Inputs and GT & Super SloMo & DAIN & SoftSplat &  Ours w/o \XXX  &  Ours \\
    \end{tabular}
        }
    \caption{\textbf{Qualitative results on  ATD-12K test set.} 
    The first row of each example shows the overlapped input frames and the optical flow predicted by corresponding method, while the second row shows the ground truth and interpolation results.
    } 
%
%
    %
      \vspace{-4pt}
    \label{fig:compare}
\end{figure*}

\subsection{Comparative Results} \label{subsect:quantitative_eval}

\noindent\textbf{Quantitative Evaluation.}
The quantitative results are shown in Table \ref{table:quantitative_eval}.
%
%
%
According to the comparison, our proposed model consistently performs favorably against all the other existing methods on all evaluations.
The PSNR score of our method improves 0.34dB compared to the best comparative method (SoftSplat) on the whole image evaluation.
For the evaluation on RoI, which indicates the significant movement in a triplet, our method can also achieve an improvement of 0.32dB.
Since animation videos contain many frames with static background, the evaluation on RoI is more precise to reflect the effectiveness on large motions.
%

\noindent\textbf{Qualitative Evaluation.}
To further demonstrate the effectiveness of our proposed method, we show two examples for visual comparison in Figure \ref{fig:compare}.
In each example, we display the interpolation results and corresponding optical flows predicted by different methods.
%
%
%
In the first example, {\it Mowgli} is reaching out a vulture using one hand, while his other hand is already on the wing of the vulture. 
Interpolation on this case is very challenging because the displacement of the boy's arm is locally discontinuous and extremely large.
Moreover, the local patterns of the boy's two hands are very similar, which result in a local minimum for the optical flow prediction.
Consequently, all existing methods fail to predict correct flow values on the boy's hand.
%
In the interpolation results of these methods, the moving hand  either disappears (\eg, SoftSplat) or splits into two deformed ones (\eg, Super SloMo, DAIN).
%
%
%
In contrast, our method can estimate a more accurate optical flow and generate a relatively complete arm in the intermediate position.
In the second case, \textit{Peter Pan} flies fast in the air while the background moves with a large displacement.
%
Similar to the first example, the compared methods estimate wrong flow values on the character,
and hence fail to synthesize a complete body in the accurate position,
while our method can produce the whole character, which is almost identical to the ground truth and looks correct.

\begin{figure*}[h]
    \scriptsize
    \setlength{\tabcolsep}{1.5pt}
    \centering
    \includegraphics[width=1\linewidth,  trim=0pt 10pt 0pt 60pt, clip ]{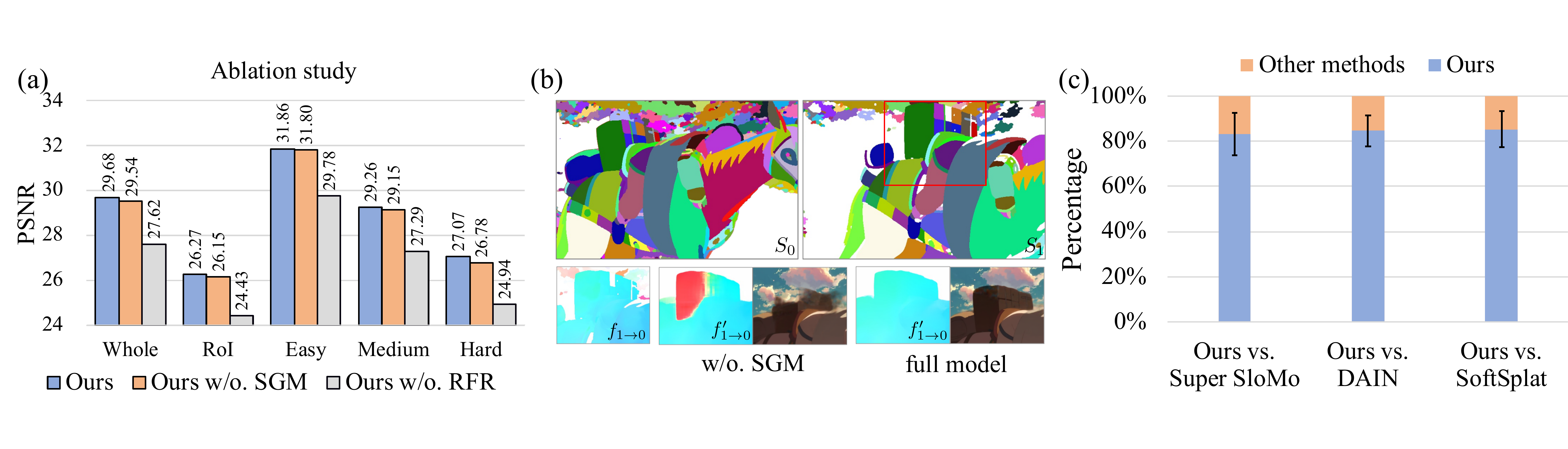}\\
    \vspace{-6pt}
    \caption{
        (a) \textbf{Ablation study for the SGM module.} Quantitative results of PSNR are exhibited. (b) \textbf{Visualization for the effectiveness of SGM.} In the first row, the color piece segmentation of frame 0 and frame 1 in Figure \ref{fig:teaser} are visualized, where the matching pieces are filled with the same random color in $S_0$ and $S_1$. In the second row, flow estimations and final interpolation results are displayed. (c) \textbf{User study results.} Our method outperforms others with a large margin. 
        }
    \label{fig:aus}
    \vspace{-6pt}
\end{figure*}

\subsection{Ablation Study}

\noindent\textbf{Quantitative Evaluation.}
%
To explain the effectiveness of the SGM and RFR modules in AnimeInterp, 
we evaluate two variants of the proposed method, where the SGM module and the RFR module are removed, respectively (denote as ``Ours w/o. SGM'' and ``Ours w/o. RFR'').
The ``w/o. SGM'' variant directly predicts optical flows using the recurrent refinement network, without the guided initialization of global-context-aware piece-wise flow, while the ``w/o. RFR'' model uses the  piece-wise flow predicted by SGM to warp and synthesize the interpolate results.
For a fair comparison, we re-fine-tune the weights of these two variants after removing the corresponding modules.
%
%
As shown in Table \ref{table:quantitative_eval} and Figure \ref{fig:aus}(a), the PSNR of `` w/o. SGM'' variant on the whole-image evaluation is lowered about 0.14dB
, while the score of ``w/o. RFR'' model drops sharply to 2.06dB, because the piece-wise flow are coarse and contain zero values on mutually inconsistent color pieces.
The results suggest that both of the proposed SGM module and RFR module play a critical role in our method.
To have a more precise view on different difficulty levels, the SGM module can improve about 0.3dB on the performance of triplets marked as ``Hard'', from 26.78dB to 27.07dB, but merely improves the ``Easy'' evaluation by 0.06dB, 
suggesting that the proposed SGM module is more useful on improving the performance of large-motion videos.

\noindent\textbf{Qualitative Evaluation.}
%
%
As for the quality of visual results, if the SGM module is removed, the predicted flow of our methods becomes incorrect since it falls into a local minimum.
For instance, in the first sample of Figure \ref{fig:compare}, the moving right hand of the boy in the first image is matched to the left hand in the second image, which is spatially closer but wrong for the matching.
%
Interestingly, with the piece-wise flow as guidance, our method can avoid this local minimum and estimate correct flow values by taking advantage of global context information.
%
%
A detailed illustration of the effectiveness of SGM is shown in Figure \ref{fig:aus}(b).
In the first row, we show the color piece segmentation of the two input frames displayed in Figure \ref{fig:teaser}.
The matching pieces are filled with the same random color, and the color pieces disobeying the mutual consistency are masked as white.
%
%
As can be seen in the red boxes, the segments of the luggage are correctly matched, even though it is segmented into several pieces.
In the second row of Figure \ref{fig:aus}(b), we visually compare the optical flow and the interpolation results with and without SGM.
Although the piece-wise flow $f_{1\to 0}$ is coarse, it can be a good guidance to the refinement network to predict accurate $f'_{1\to0}$, which leads to a complete synthesis of the luggage.
In conclusion, the proposed SGM module significantly improves the performance on local-discontinuous motions via global context matching.

\subsection{Further Analysis}
%
%

\noindent\textbf{Influence of Difficulty Levels and Motion RoIs.}
As can be seen in Table \ref{table:quantitative_eval}, the difficulty levels have great impact on the performance.
The PSNR scores of various methods will drop more than 2dB when the difficulty increases one level.
The drop is more significant for those methods which are not designed to resolve ``lack of textures'' and ``non-linear and extremely large motion'' challenges in animation.
%
For example, DAIN achieves the third-best score (31.67dB) in the evaluation on `Easy' level, with the difference less than 0.2dB compared to our proposed method (31.86dB), but its performance drops more than ours for `Medium' (28.74dB vs. 29.26dB) and `Hard' (26.22dB vs. 27.07dB) levels.
%
Meanwhile, the performance of various methods on RoIs are much lower than that on the whole images. 
Since RoI regions are the essential parts of animation videos, which have great impact on the visual experience, future studies focusing on restoring RoIs are encouraged.

\noindent\textbf{User Study.}
To further evaluate the visual performance of our methods, we conduct a user study on the interpolation results of Super SloMo, DAIN, SoftSplat and our proposed method. 
%
We separately conduct subjective experiments with ten people.
In every test, we randomly select 150 pairs, each of which contains the interpolation result of our method and the corresponding frame from one of the compared methods, and ask the subject to vote for the better one.
%
We provide both the interpolated frames and the input frames to subjects so that they can take the temporal consistency in their decision.
Figure \ref{fig:aus} shows the percentages of the average votes of our method versus the compared methods.
For each compared method, over 83\% of the average votes account for ours as being better in terms of visual quality.
These experimental results show our method produces more favorable results on both the quantitative evaluations and the visual quality.
\vspace{-2pt}

%% file: content/conclusion.tex
This paper introduces the animation video interpolation task, which is an important step towards cartoon video generation. To benchmark animation video interpolation, we build a large-scale animation triplet dataset named ATD-12K comprising 12,000 triplets with rich annotations.  We also propose an effective animation video interpolation framework called AnimeInterp, which employs the segmentation of color pieces as a match guidance to compute piece-wise optical flows between different frames and leverages a recurrent module to further refine the quality of optical flows. Comprehensive experiments demonstrate the effectiveness of the proposed modules as well as the generalization ability of AnimeInterp.